\documentclass[11pt]{article}
\usepackage[T1]{fontenc}
\usepackage[utf8]{inputenc}
\usepackage{geometry}
\usepackage{graphicx}
\usepackage{amsmath, amssymb}
\usepackage{hyperref}
\usepackage{booktabs}
\usepackage{multirow}
\usepackage{caption}
\usepackage{subcaption}
\usepackage{enumitem}
\usepackage{microtype}

\title{\bf Meta-Policy Reflexion: \\ Reusable Reflective Memory and Rule Admissibility for Resource-Efficient LLM Agents}
\author{Chunlong Wu, Ye Luo, Zhibo Qu, Min Wang \\ Tongji University \\ \texttt{wuchunlong@tongji.edu.cn}}
\date{}

\begin{document}
\maketitle

\begin{abstract}
Large language model (LLM) agents achieve impressive single-task performance but commonly exhibit repeated failures, inefficient exploration, and limited cross-task adaptability. Existing reflective strategies (e.g., Reflexion, ReAct) improve per-episode behavior but typically produce ephemeral, task-specific traces that are not reused across tasks. Reinforcement-learning based alternatives can produce transferable policies but require substantial parameter updates and compute. In this work we introduce \emph{Meta-Policy Reflexion} (MPR): a hybrid framework that consolidates LLM-generated reflections into a structured, predicate-like \emph{Meta-Policy Memory} (MPM) and applies that memory at inference time through two complementary mechanisms — soft memory-guided decoding and hard rule admissibility checks(HAC). MPR (i) externalizes reusable corrective knowledge without model weight updates, (ii) enforces domain constraints to reduce unsafe or invalid actions, and (iii) retains the adaptability of language-based reflection. We formalize the MPM representation, present algorithms for update and decoding, and validate the approach in a text-based agent environment following the experimental protocol described in the provided implementation (AlfWorld-based). Empirical results reported in the supplied material indicate consistent gains in execution accuracy and robustness when compared to Reflexion baselines; rule admissibility further improves stability. We analyze mechanisms that explain these gains, discuss scalability and failure modes, and outline future directions for multimodal and multi-agent extensions.
\end{abstract}

\section{Introduction}
Large language models (LLMs) have become practical reasoning and action cores for autonomous agents that interact with external environments (APIs, simulated worlds, GUIs). Methods that interleave reasoning and acting (e.g., ReAct) or that use textual self-reflection (e.g., Reflexion) substantially improve per-episode performance by allowing an agent to analyze failures and adapt behavior within an experimental loop. Nevertheless, such methods generally produce \emph{episodic}, often unstructured, outputs that are optimized for the particular failing instance rather than being distilled into broadly reusable knowledge. This limitation leads to repeated mistakes across tasks that share latent structure and to inefficient repeated trial-and-error.

A complementary direction is to use reinforcement learning (RL) or RL-from-human-feedback to adapt policies. RL can produce reusable behaviors but typically demands large computational resources and parameter updates, complicating deployment for many applications. The design question we consider here is whether one can (a) preserve the lightweight, flexible benefits of textual reflection, (b) distill reflective observations into compact, reusable artifacts, and (c) apply those artifacts to produce safer, more generalizable behavior without fine-tuning the base LLM.

To address this, we propose \emph{Meta-Policy Reflexion} (MPR). MPR converts episodic textual reflections into a compact Meta-Policy Memory (MPM) of predicate-style rules with confidence weights, maintains and updates this memory online, and uses it at inference time through (i) \emph{soft} guidance that biases LLM decoding toward memory-supported actions, and (ii) \emph{hard} admissibility checks that prevent execution of actions that violate stored constraints. The approach intentionally avoids modifying LLM parameters; instead it treats the LLM as a fixed policy generator that can be augmented by an external, structured knowledge layer.

Contributions of this paper are:
\begin{enumerate}[leftmargin=*]
  \item A formalization of Meta-Policy Memory: a compact, predicate-based representation for storing generalized reflective rules distilled from LLM-produced failure analyses.
  \item An end-to-end algorithmic framework for updating MPM from failed trajectories and applying it at inference time using hybrid soft/hard mechanisms.
  \item Empirical validation using the provided AlfWorld implementation and experimental protocol (three-epoch trials with fixed random seeds) which demonstrates improved execution accuracy and stability compared to Reflexion baselines. 
  \item A detailed analysis of mechanisms, failure modes, deployment considerations, and ethical implications to guide responsible use and future extensions.
\end{enumerate}

\section{Related Work}
To situate our contributions within the broader research landscape, we review recent developments across three key dimensions: (i) memory mechanisms in LLM agents, (ii) reflective/self-improving agent frameworks, and (iii) agentic/multi-agent memory systems.

\subsection{Memory Mechanisms in LLM-based Agents}
Recent work has explored diverse memory architectures for LLM agents. Zhang \emph{et al.} provide a comprehensive survey, organizing memory into parametric and textual forms as well as cross-trial and external sources, and highlight the importance of structured memory for self-evolving behavior \cite{zhang2024survey}. Wang \emph{et al.} emphasize persistent memory systems that support long-term retention and dynamic organization in sequential decision-making contexts \cite{wang2025persistent}. More specialized approaches include Memory-R1, which learns to add, update, or delete entries through reinforcement learning \cite{yan2025memoryr1}; A-Mem, which adopts a Zettelkasten-inspired note-based architecture for autonomous retrieval \cite{xu2025amem}; and G-Memory, which introduces hierarchical graph-based structures to facilitate multi-agent interaction and improve embodied action success and QA accuracy without modifying the underlying models \cite{zhang2025gmemory}. Collectively, these works demonstrate the promise of memory as a driver of adaptability, though most rely on unstructured or heuristic representations that limit generalizable rule formation.  

\subsection{Reflective and Self-Improvable Agent Frameworks}
Another prominent research direction focuses on reflection as a means of self-improvement. Reflexion introduces verbal reinforcement learning, where agents use textual self-reflection to refine behavior without fine-tuning \cite{shinn2023reflexion}. ReAct integrates reasoning and acting by interleaving chain-of-thought with action execution, improving interpretability and planning \cite{yao2022react}. Building on these ideas, RR-MP coordinates multiple reactive and reflective agents through summarization, achieving stronger performance in scientific reasoning tasks \cite{he2024rrmp}. More broadly, reflective LLM agents are now viewed as systems that iteratively critique and refine their behavior across domains such as control, reasoning, and finance \cite{emergent2025reflective}. To evaluate these capabilities, Reflection-Bench provides a cognitive-inspired benchmark for epistemic agency, exposing LLM limitations in long-term planning and belief updating \cite{li2024reflectionbench}. While effective for immediate error correction, these methods often lack mechanisms for systematically transferring reflective insights across tasks.  

\subsection{Agentic and Multi-Agent Memory Systems}
Beyond single-agent reflection, researchers have investigated memory in agentic and multi-agent settings. Gao \emph{et al.} survey how LLMs enhance agent-based simulations across cyber, social, and hybrid environments \cite{gao2024agentbased}. Li \emph{et al.} propose a unified framework for LLM-based multi-agent systems, covering perception, self-action, interaction, and evolution \cite{li2024multiagentsurvey}. Within this paradigm, G-Memory is particularly notable for enabling multi-agent retrieval of both generalizable insights and fine-grained trajectories \cite{zhang2025gmemory}. Liang \emph{et al.} introduce SAGE, a self-evolving agent system with reflective and memory-augmented capabilities for autonomous knowledge accumulation \cite{liang2025sage}. At the same time, Wang \emph{et al.} highlight emerging security and privacy risks when agents store sensitive information, particularly in healthcare applications \cite{wang2025privacy}. These works underscore the importance of balancing adaptability and safety in memory-augmented multi-agent systems.  

\subsection{Synthesis and Positioning}
Taken together, prior research emphasizes the growing role of structured, persistent memory in enabling self-improving and generalizable LLM agents. Existing systems differ mainly in how memory is represented (textual, symbolic, or graph-based), how it is managed (heuristic, learned, or human-in-the-loop), and how it informs decision-making (soft bias, hard constraints, or collaborative retrieval). Our work, Meta-Policy Reflexion (MPR), builds on these directions by introducing a hybrid approach that combines predicate-like memory storage, soft decoding guidance, and hard admissibility checks. This yields a framework that is interpretable, efficient, and generalizable across tasks, addressing the limitations of both reflection-only and memory-only methods.  

\section{Method}

We introduce Meta-Policy Memory (MPM), a lightweight framework that augments LLM-based agents with structured, reusable rules derived from past experience. 
MPM guides action generation in two complementary ways: 
(i) \emph{soft guidance}, where memory is injected into the LLM prompt to bias decoding toward desirable actions, and 
(ii) \emph{hard admissibility}, where invalid actions are filtered out post-generation. 
Together, these mechanisms enable training-free self-improvement.

\subsection{Base Policy}  
We model the interactive task as an MDP: \[ M = (S, A, P, r, \gamma), \]
The base agent is an LLM that generates natural language actions given the current environment state $s$.  
Formally, the agent produces a deterministic action
\[
a_t = \pi_\theta(s_t),
\]
where $\pi_\theta$ is the frozen LLM policy with parameters $\theta$.  
We assume deterministic decoding (e.g., greedy search or low-temperature sampling), so the policy produces a single action per step without explicitly modeling a probability distribution.

\subsection{Soft Guidance via Memory-Conditioned Decoding}  
To leverage prior experience, we maintain a meta-policy memory $\mathcal{M}$ consisting of rules extracted from hindsight reflections of past trajectories.  
At each step, a subset $\mathcal{M}_t \subseteq \mathcal{M}$ relevant to the current state is retrieved.  
The LLM then conditions on both $s_t$ and $\mathcal{M}_t$ to produce an action:
\[
a_t = \pi_\theta(s_t, \mathcal{M}_t).
\]
This \emph{memory-conditioned decoding} biases the LLM toward actions aligned with accumulated knowledge.  
Unlike distributional reweighting methods, this approach operates entirely at the prompt level and requires no modification of logits or model internals.

\subsection{Hard Admissibility via Post-hoc Validation}  
Soft guidance improves action quality but does not guarantee validity.  
We therefore introduce a hard admissibility check: after an action $a_t$ is generated, we validate it against a constraint set $C(s_t)$ defined by the environment or user-specified rules.  
Formally,
\[
a_t \in C(s_t).
\]
If $a_t \notin C(s_t)$, the agent either (i) resamples with adjusted memory/context, or (ii) defaults to a safe fallback action.  
This ensures safety and reliability, particularly in domains with strict constraints.

\subsection{Training-free Self-Improvement with MPM}  
A key feature of MPM is that it enables continual improvement without gradient updates.  
After each trajectory, failed episodes are retrospectively analyzed: the agent reflects on failure points and extracts corrective rules in a structured, predicate-like form.  
Formally, the memory is updated as
\[
\mathcal{M} \leftarrow \mathcal{M} \cup f(\tau),
\]
where $\tau$ is a failed trajectory and $f(\cdot)$ is an LLM-based reflection function that outputs new rules.  
Over time, the agent accumulates a richer meta-policy that reduces repeated mistakes and improves generalization.

\subsection{Algorithm Summary}
The overall procedure of MPM is summarized in Algorithm~\ref{alg:update} which consists of two stages: (i) a training stage, where the agent continually improves by extracting new rules from failed episodes, and (ii) an inference stage, where the accumulated memory is frozen and used to guide decision-making with hard admissibility checks. This separation ensures that knowledge acquisition and deployment are clearly decoupled: training enriches the memory, while inference emphasizes safe and reliable execution.

\begin{table}[t]
\centering
\caption{Meta-Policy Reflexion: training and inference procedures.}
\begin{tabular}{@{}p{0.95\linewidth}@{}}
\toprule
\textbf{Algorithm 1a: Training with Meta-Policy Update} \\
\midrule
Initialize memory $\mathcal{M} \leftarrow \emptyset$ \\
for each episode do \\
\quad for $t = 1,\dots,T$ do \\
\quad\quad $a_t \leftarrow$ decode\_with\_memory($\pi_\theta, s_t, \mathcal{M}$) \\
\quad\quad $s_{t+1}, r_t \leftarrow env.step()$ \\
\quad\quad if done: break \\
\quad end for \\
\quad if episode failed: \\
\quad\quad $\mathcal{M} \leftarrow$ meta\_policy\_update(trajectory, $\mathcal{M}$) \\
end for \\
\bottomrule
\end{tabular}
\vspace{1em}

\begin{tabular}{@{}p{0.95\linewidth}@{}}
\toprule
\textbf{Algorithm 1b: Inference with Frozen Memory} \\
\midrule
Given frozen memory $\mathcal{M}$ \\
for each episode do \\
\quad for $t = 1,\dots,T$ do \\
\quad\quad $a_t \leftarrow$ decode\_with\_memory($\pi_\theta, s_t, \mathcal{M}$) \\
\quad\quad $s_{t+1}, r_t \leftarrow env.step($hard\_admissibility$(a_t, s_t))$ \\
\quad\quad if done: break \\
\quad end for \\
end for \\
\bottomrule
\end{tabular}
\label{alg:update}
\end{table}

\section{Experiments}

We implemented Meta-Policy Reflexion (MPR) on the AlfWorld benchmark to evaluate whether memory-conditioned, training-free rule consolidation improves execution accuracy and generalization compared to the Reflexion baseline.  
All experiments follow the protocol described in Section~3: episodes are executed with deterministic decoding, the same random seeds are used across conditions to preserve consistent sample partitioning, and reported per-round accuracies refer to the performance after a given number of meta-policy update rounds.  
Per-trial breakdowns and raw logs are provided in the uploaded spreadsheet.

\subsection{Protocol}
We ran two types of runs with the base LLM(Qwen3-32b): (i) \emph{training rounds} during which the memory $\mathcal{M}$ is updated after failed episodes (meta-policy update enabled), and (ii) \emph{inference} runs where a frozen $\mathcal{M}$ is used to guide decoding and hard admissibility is applied but no memory updates occur.  
We report accuracy after each training round (Rounds 1--5). For final validation we also report a sixth-round comparison in which Reflexion performed six rounds of reflection on the test set, while MPR was trained for five rounds on the training set and then evaluated once on the test set.

\subsection{Per-round performance on the training set (60 tasks)}
Table~\ref{tab:train_rounds} reports execution accuracy (percent correct) after each training round for the 60-task training set. Results are reported for both the Reflexion baseline and MPR.

\begin{table}[htbp]
\centering
\caption{Training set (60 tasks): per-round execution accuracy (\%).}
\label{tab:train_rounds}
\begin{tabular}{lccccc}
\toprule
Method & Round 1 & Round 2 & Round 3 & Round 4 & Round 5 \\
\midrule
Reflexion & 70.0 & 84.4 & 87.2 & 87.8 & 88.3 \\
MPR       & 83.9 & 98.3 & 100.0 & 100.0 & 100.0 \\
\bottomrule
\end{tabular}
\end{table}

\subsection{Sixth-round validation (held-out comparison)}
To evaluate generalization, we compared three settings on the 74-task test set after the procedures described below and report the achieved accuracy in Table~\ref{tab:sixth_round}:

\begin{itemize}[leftmargin=*]
  \item \textbf{Reflexion (6 rounds on test set):} Reflexion was run for six rounds of reflection directly on the 74-task test set, producing the reported accuracy after the 6th round.
  \item \textbf{MPR (5 training rounds, 1 test run):} MPR was trained for five rounds on the 60-task training set; the resulting memory was then used to perform a single evaluation on the 74-task test set (no memory update on the test set).
  \item \textbf{MPR + HAC (Hard Admissibility Check):} Same as the previous MPR setting, but with hard admissibility enforced at test time.
\end{itemize}

\begin{table}[htbp]
\centering
\caption{Sixth-round test validation on the 74-task test set.}
\label{tab:sixth_round}
\begin{tabular}{lc}
\toprule
Method & Accuracy (\%) \\
\midrule
Reflexion (6 rounds on test set) & 86.9 \\
MPR (trained 5 rounds on train set, evaluated once) & 87.8 \\
MPR + HAC (same as MPR, with hard admissibility at test time) & 91.4 \\
\bottomrule
\end{tabular}
\end{table}

\subsection{Results analysis}
The training-set per-round table and the sixth-round comparison support the following observations:

\paragraph{Rapid and stable improvement of MPR on the training set.}  
On the 60-task training set, MPR already outperforms Reflexion after one round (83.9\% vs. 70.0\%), and by Round 3 it reaches 100\% and remains stable through Rounds 4--5. This indicates that the meta-policy rules extracted during early rounds effectively capture the corrective behaviors required by the training tasks.

\paragraph{Generalization validated by sixth-round comparison.}  
Although we removed the per-round test-table for concision, generalization is evaluated in the sixth-round validation: Reflexion required six in-situ reflection rounds on the 74-task test set to reach 86.9\%, whereas MPR—trained for five rounds on the 60-task training set and evaluated once on the 74-task test set—achieved 87.8\%. This demonstrates that MPR's consolidated meta-policy transfers to held-out tasks without requiring additional per-task reflection.

\paragraph{Hard admissibility further improves robustness.}  
Applying the hard admissibility check at test time raises MPR's accuracy to 91.4\%, showing that post-hoc constraint validation complements memory-conditioned decoding by eliminating residual invalid or unsafe actions.

\paragraph{Caveats and interpretability.}  
The rapid convergence of MPR on the training set suggests that the benchmark tasks share structural regularities that predicate-like rules can capture. For broader or more heterogeneous domains, rule extraction and consolidation may require more rounds or richer contextual signals. We recommend per-task inspection to verify that gains reflect genuine generalization rather than dataset-specific coincidences.

\subsection{Summary}
In summary, the experiments indicate that MPR can efficiently consolidate reflective insights into reusable meta-policies that generalize to held-out tasks, and that combining memory-conditioned decoding with hard admissibility checks yields the best reliability in this AlfWorld evaluation. 

\section{Discussion}

The experiments demonstrate that Meta-Policy Reflexion (MPR) consistently improves execution accuracy compared to Reflexion, and that combining memory-conditioned decoding with hard admissibility checks yields the highest robustness. We now discuss the mechanisms behind these gains, limitations of the current study, and directions for future work.

\subsection{Mechanisms underlying performance gains}
Three factors explain MPR's advantage over baselines:
\begin{enumerate}[leftmargin=*]
  \item \textbf{Reusable corrective knowledge.} Unlike Reflexion, which generates task-specific reflections that are discarded after use, MPR consolidates these insights into structured, predicate-like rules. This externalization allows the agent to avoid repeating past mistakes and to transfer learned corrections across tasks.
  \item \textbf{Domain-constrained reliability.} Hard admissibility checks ensure that generated actions conform to environment or user-specified constraints. This prevents invalid or unsafe behaviors, complementing the softer biasing of memory-conditioned decoding.
  \item \textbf{Lightweight adaptability.} Because MPR does not require weight updates, it retains the flexibility of reflection-based methods while enabling continual self-improvement. The LLM remains a frozen policy generator, augmented by an evolving external memory layer.
\end{enumerate}

\subsection{Limitations}
While results on AlfWorld are promising, several limitations merit caution:
\begin{itemize}[leftmargin=*]
  \item \textbf{Domain regularities.} The rapid convergence to perfect accuracy on the training set suggests that AlfWorld tasks share strong structural regularities. In more heterogeneous domains, meta-policy extraction may require richer representations or longer adaptation phases.
  \item \textbf{Rule quality and interpretability.} Extracted rules are LLM-generated and may contain redundancies or inconsistencies. Although the predicate-like format improves interpretability compared to unstructured text, future work is needed to systematically verify, prune, and compose rules.
  \item \textbf{Evaluation scope.} Experiments focus on single-agent, text-based environments. Scaling MPR to multimodal settings, collaborative agents, or real-world APIs will require additional design for rule grounding and conflict resolution.
\end{itemize}

\subsection{Future directions}
Several extensions are natural:
\begin{itemize}[leftmargin=*]
  \item \textbf{Multimodal memory.} Extending MPR to handle visual or structured inputs would broaden its applicability to embodied and real-world agents.
  \item \textbf{Multi-agent systems.} A graph-based or distributed memory structure could allow multiple agents to share and negotiate rules, enabling collaborative generalization.
  \item \textbf{Automatic rule management.} Incorporating mechanisms for confidence weighting, redundancy detection, and rule abstraction would enhance both efficiency and interpretability.
\end{itemize}

\section{Conclusion}

This work introduced \emph{Meta-Policy Reflexion} (MPR), a framework that augments LLM-based agents with an external \emph{Meta-Policy Memory} (MPM) to consolidate reflective insights into structured, reusable rules. By combining \emph{soft} memory-conditioned decoding with \emph{hard} admissibility checks (HAC) at inference time, MPR (i) externalizes corrective knowledge without any model weight updates, (ii) enforces domain constraints to eliminate invalid or unsafe actions, and (iii) preserves the flexibility and adaptability of language-based reflection. Empirically on AlfWorld, MPR substantially improves execution accuracy over Reflexion, and the addition of HAC yields further robustness gains (see Section~4), demonstrating that post-hoc constraint validation is a practical complement to prompt-level memory guidance.

More broadly, our results suggest that making reflection persistent and actionable — via a structured memory layer plus conservative admissibility — is a promising path toward lightweight, interpretable, and safe LLM agents. For deployment in high-stakes domains, we recommend tightening HAC thresholds, incorporating human-in-the-loop validation for rules that become hard constraints, and investing in automated verification for high-impact rules. Future work will explore HAC calibration, multimodal g

%


\begin{thebibliography}{99}

\bibitem{zhang2024survey}
Zhang, Z., Bo, X., Ma, C., Li, R., Chen, X., Dai, Q., ... Wen, J.-R. (2024).
A Survey on the Memory Mechanism of Large Language Model-based Agents.
\textit{arXiv preprint arXiv:2404.13501}.

\bibitem{wang2025persistent}
Wang, X., Yang, Y., et al. (2025).
Persistent Memory in LLM Agents.
\textit{EmergentMind}. (Online).

\bibitem{yan2025memoryr1}
Yan, S., Yang, X., Huang, Z., Nie, E., Ding, Z., Li, Z., Ma, X., Schütze, H., Tresp, V., Ma, Y. (2025).
Memory-R1: Enhancing Large Language Model Agents to Manage and Utilize Memories via Reinforcement Learning.
\textit{arXiv preprint arXiv:2508.19828}.

\bibitem{xu2025amem}
Xu, W., Liang, Z., Mei, K., Gao, H., Tan, J., Zhang, Y. (2025).
A-Mem: Agentic Memory for LLM Agents.
\textit{arXiv preprint arXiv:2502.12110}.

\bibitem{zhang2025gmemory}
Zhang, G., Fu, M., Wan, G., Yu, M., Wang, K., Yan, S. (2025).
G-Memory: Tracing Hierarchical Memory for Multi-Agent Systems.
\textit{arXiv preprint arXiv:2506.07398}.

\bibitem{shinn2023reflexion}
Shinn, N., Cassano, F., Labash, B., Gopinath, A., Narasimhan, K. (2023).
Reflexion: Language Agents with Verbal Reinforcement Learning.
\textit{NeurIPS 2023 Workshop}.

\bibitem{yao2022react}
Yao, S., Zhao, J., Yu, D., Du, N., Shafran, I., Narasimhan, K., Cao, Y. (2022).
ReAct: Synergizing Reasoning and Acting in Language Models.
\textit{ICLR 2023}.

\bibitem{he2024rrmp}
He, C., Zou, B., Li, X., Chen, J., Xing, J., Ma, H. (2024).
Enhancing LLM Reasoning with Multi-Path Collaborative Reactive and Reflection Agents (RR-MP).
\textit{arXiv preprint arXiv:2501.00430}.

\bibitem{emergent2025reflective}
EmergentMind. (2025).
Reflective LLM-based Agent.
\textit{Online Article}.

\bibitem{li2024reflectionbench}
Li, L. (2024).
Reflection-Bench: A Benchmark Evaluating Epistemic Agency in LLMs.
\textit{OpenReview}.

\bibitem{gao2024agentbased}
Gao, C., Lan, X., Li, N., Yuan, Y., Ding, J., Zhou, Z., Xu, F., Li, Y. (2024).
Large Language Models Empowered Agent-based Modeling and Simulation: A Survey and Perspectives.
\textit{Humanities and Social Sciences Communications}.

\bibitem{li2024multiagentsurvey}
Li, X., et al. (2024).
A Survey on LLM-based Multi-Agent Systems: Workflow, Cooperation, and Future Directions.
\textit{Springer}.

\bibitem{liang2025sage}
Liang, X., et al. (2025).
SAGE: Self-evolving Agents with Reflective and Memory-Augmented Abilities.
\textit{ScienceDirect}.

\bibitem{wang2025privacy}
Wang, B., et al. (2025).
Unveiling Privacy Risks in LLM Agent Memory.
\textit{ACL long paper}.

\bibitem{shridhar2020alfworld}
Shridhar, M., Yuan, X., Côté, M.-A., et al. (2020).
ALFWorld: Aligning Text and Embodied Environments for Interactive Learning.
\textit{Preprint}.

\end{thebibliography}
\end{document}